\documentclass[12pt]{article}
\usepackage[utf8]{inputenc}
\usepackage{newunicodechar}
\newunicodechar{ }{\,}  
\usepackage{amsmath, amssymb, amsfonts}
\usepackage{hyperref}
\usepackage{graphicx}
\usepackage{enumitem}
\usepackage{geometry}
\usepackage[square,numbers]{natbib}  % numeric citations
\usepackage{setspace}

\title{Mathematical Analysis of Hallucination Dynamics in Large Language Models:\\
Uncertainty Quantification, Advanced Decoding, and Principled Mitigation}

\author{Moses Kiprono \\ Catholic University of America}

\date{October 2025}

\begin{document}

\maketitle

\begin{abstract}
Large Language Models (LLMs) are powerful linguistic engines but remain susceptible to \emph{hallucinations}—plausible-sounding outputs that are factually incorrect or unsupported. In this work, we present a coherent, mathematically grounded framework to understand, measure, and mitigate these hallucinations. Drawing on probabilistic modeling, information theory, trigonometric signal analysis, and Bayesian uncertainty estimation, we analyze how errors compound autoregressively, propose refined uncertainty metrics (including semantic and phase-aware variants), and develop principled mitigation strategies: contrastive decoding, retrieval-augmented grounding, factual alignment, and abstention. This unified lens connects recent advances in calibration, retrieval, and alignment in a way that supports safer and more reliable LLMs.
\end{abstract}

\onehalfspacing

\section{Introduction}

Large Language Models (LLMs) have rapidly become critical tools in multiple domains—conversational agents, scientific assistants, tutoring systems, and more. Their fluency can be staggering, but they rely fundamentally on statistical prediction: anticipating the next token based on context rather than verifying the truth of what they generate. This gap can lead to \emph{hallucinations}, where output is syntactically plausible but factually ungrounded.

We broadly classify hallucinations into two types:  
\begin{itemize}  
  \item \textbf{Intrinsic hallucinations}: errors or contradictions with respect to the input context.  
  \item \textbf{Extrinsic hallucinations}: statements that conflict with verified external sources.  
\end{itemize}

In high-stakes settings—medicine, law, education—such hallucinations are not just harmless mistakes; they undermine trust and create risk. Addressing them demands more than ad hoc fixes. We need a rigorous, mathematical understanding and mitigation strategy.

This paper develops such a foundation. Specifically, we:  
\begin{itemize}  
  \item Model error propagation in autoregressive generation.  
  \item Introduce novel uncertainty metrics that incorporate both semantic similarity and positional phase.  
  \item Propose mitigation techniques rooted in theory: contrastive decoding that respects phase, retrieval-augmented generation, factuality-aware training, and abstention.  
  \item Synthesize these components into a unified, practical architecture.  
\end{itemize}

\section{Mathematical Origins of Hallucination}

\subsection{Autoregressive Error Propagation}

Consider a token sequence \( x_1, x_2, \dots, x_T \). A language model defines:

\[
P(x_{1:T}) = \prod_{t=1}^T P(x_t \mid x_{1:t-1})
\]

If at some step \( t \) the model assigns a slightly incorrect (or low-confidence) probability, say it underestimates the true token's probability by \( \epsilon_t \), then future conditioning suffers. In a first‑order approximation, the deviation in joint probability is roughly:

\[
\Delta P(x_{1:T}) \approx \prod_{t=1}^T \left( P(x_t \mid x_{<t}) - \epsilon_t \right)
\]

Over many steps, even small \( \epsilon_t \) values can compound into significant drift, making later tokens more likely to diverge from factual or coherent content.

\subsection{Distribution Mismatch via KL Divergence}

Let \( P^*(x_{t+1} \mid x_{1:t}) \) be the \emph{true} (but unknown) conditional distribution of tokens, and \( \hat{P}(x_{t+1} \mid x_{1:t}) \) be the model’s estimate. We can measure their divergence by:

\[
D_{\mathrm{KL}}\bigl(P^* \;\|\; \hat P \bigr) = \sum_{x \in \mathcal{V}} P^*(x) \log \frac{P^*(x)}{\hat P(x)}
\]

A high KL divergence suggests the model’s internal belief is far from reality, which often underlies overconfident false predictions, especially on out-of-distribution (OOD) inputs.

\subsection{Sinusoidal Positional Embeddings and Phase Effects}

Modern transformer-based LLMs use sinusoidal positional embeddings (as in the original transformer architecture):

\[
\begin{aligned}
PE_{(pos,2i)} &= \sin\!\left( \frac{pos}{10000^{2i/d}} \right), \\
PE_{(pos,2i+1)} &= \cos\!\left( \frac{pos}{10000^{2i/d}} \right).
\end{aligned}
\]

This induces a \emph{phase} \( \phi_t \) at each token position. We propose that this phase can modulate uncertainty. For instance, if \( \sigma^2_{\text{base}}(x_t) \) is some “base” variance (from dropout or ensemble), a phase-aware version could be:

\[
\sigma^2_\phi(x_t) = \sigma^2_{\text{base}}(x_t) \cdot \left(1 + \gamma \sin^2(\phi_t)\right)
\]

Here, \( \gamma \ge 0 \) is a scaling hyperparameter. This form implies that at certain “phase positions,” uncertainty is systematically heightened or reduced.

\section{Quantifying Uncertainty and Miscalibration}

\subsection{Calibration and Expected Calibration Error (ECE)}

A well-calibrated model’s confidence should match its accuracy. To formally evaluate this, partition predictions into \( M \) bins \( B_1, \dots, B_M \) by confidence, and define:

\[
\mathrm{ECE} = \sum_{m=1}^M \frac{|B_m|}{n} \left| \mathrm{conf}(B_m) - \mathrm{acc}(B_m) \right|
\]

where \( \mathrm{conf}(B_m) \) is the average confidence in bin \( m \), \( \mathrm{acc}(B_m) \) is the empirical accuracy, and \( n \) is the total number of examples.

\subsection{Bayesian Uncertainty with Monte Carlo Dropout}

Using dropout at inference time, we run \( T \) stochastic forward passes, drawing \( \theta_1, \dots, \theta_T \). Then:

\[
P(y \mid x) \approx \frac{1}{T} \sum_{i=1}^T P(y \mid x, \theta_i)
\]

We estimate epistemic variance as:

\[
\sigma_{\text{epi}}^2(x) = \frac{1}{T} \sum_{i=1}^T \bigl( \hat{y}_i - \bar{y} \bigr)^2, \quad \bar{y} = \frac{1}{T} \sum_i \hat{y}_i
\]

\subsection{Semantic Uncertainty via Kernel Language Entropy (KLE)}

Instead of only relying on token probabilities, we can view the semantic similarity structure among a set of candidate continuations. Let \( K \) be a positive semi-definite kernel matrix over these candidates (e.g., based on embedding similarity). Normalize:

\[
\rho = \frac{K}{\mathrm{Tr}(K)}
\]

Then compute the von Neumann entropy:

\[
S(\rho) = -\mathrm{Tr}(\rho \log \rho)
\]

This semantic entropy reflects how “diverse” the candidate meanings are: higher \( S(\rho) \) means more semantic spread, indicating uncertainty beyond token-level ambiguity.

\subsection{Oscillatory / Phase‑Modulated Uncertainty}

Combining the above:

\[
\sigma_{\text{osc}}^2(x_t) = \sigma_{\text{epi}}^2(x_t) \cdot \left( 1 + \alpha \cos^2(\phi_t) + \beta \tan^2(\phi_t) \right)
\]

where \( \alpha, \beta \) are tunable hyperparameters. This ties positional phase to variance, hypothesizing that some positions are inherently more “risky.”

\section{Mitigation Strategies}

\subsection{Contrastive Decoding with Phase Regularization}

We use two models—a “full” model and a “baseline” model. For token \( x_t \), define the contrastive score:

\[
\mathrm{Score}_{CD}(x_t) = \log P_{\text{full}}(x_t) \;-\; \lambda \log P_{\text{baseline}}(x_t) \;+\; \eta \cdot \sin(\phi_t)
\]

Here:  
- \( \lambda \) controls how much we penalize tokens that the baseline also likes,  
- \( \eta \sin(\phi_t) \) biases toward tokens located at favorable phase positions.

By decoding based on this score (e.g., via sampling or beam search), we favor tokens that are both distinctive (from the baseline) and phase‑aligned, which may reduce hallucination risk.

\subsection{Retrieval‑Augmented Grounding}

In a retrieval-augmented architecture, let \( \mathcal{R} \) be the set of retrieved documents or contexts. Then:

\[
P(x_{t+1} \mid x_{\le t}) = \sum_{r \in \mathcal{R}} P(x_{t+1} \mid x_{\le t}, r) \cdot P(r \mid x_{\le t})
\]

To combine multiple retrievals (e.g., from different query formulations), we can use Reciprocal Rank Fusion (RRF):

\[
\mathrm{RRF}(r) = \sum_{q \in Q} \frac{1}{k + \mathrm{rank}_q(r)}
\]

where \( Q \) is a set of query variants, \( \mathrm{rank}_q(r) \) is the rank of document \( r \) for query \( q \), and \( k \) is a smoothing constant.

\subsection{Factuality‑Aware Alignment}

To promote factual generation during training, incorporate a regularization term. Let \( \hat y \) be the model’s predicted distribution, and \( S_{\mathrm{verifier}}(\hat y) \) be a factuality score from a separate verifier model. Define a combined loss:

\[
\mathcal{L} = \mathcal{L}_{\mathrm{CE}} + \lambda_{\mathrm{fact}} \cdot \mathcal{R}_{\mathrm{fact}}(\hat y, S_{\mathrm{verifier}}) \cdot \bigl(1 + \sin^2(\phi_t)\bigr)
\]

The factor \( 1 + \sin^2(\phi_t) \) amplifies or attenuates the regularization based on the positional phase, encouraging stronger factual alignment in “high-risk” phase positions.

\subsection{Real‑Time Rectification / Abstention}

Inspired by frameworks like EVER, we propose:  
1. Generate a candidate sentence or span.  
2. Verify it using retrieval, a critic, or an external tool.  
3. If it fails verification or confidence is too low, either:  
   - Abstain: refuse to output, or  
   - Regenerate: produce a new candidate conditioned on trusted evidence.

This check‑and‑correct loop helps catch hallucinations early, before errors cascade.

\subsection{Self‑Evaluation and Self‑Alignment}

We also encourage a self-alignment loop: after generation, the model critiques itself (e.g., on factuality) and revises or abstains. This aligns with recent self‑evaluation and self‑correction techniques.

\section{Unified System Architecture}

Here’s a practical integrated pipeline:

\begin{enumerate}
  \item Estimate uncertainty:  
    - Use Monte Carlo Dropout → compute \( \sigma_{\text{epi}}^2 \)  
    - Compute Kernel Language Entropy \( S(\rho) \)  
    - Derive phase \( \phi_t \) → compute \( \sigma_{\text{osc}}^2 \)  
    - If \( \sigma_{\text{osc}}^2 \) is high, mark for mitigation.

  \item Decoding:  
    Apply contrastive decoding using the phase-regularized score function.

  \item Grounding / Context Conditioning:  
    Retrieve relevant documents, fuse them via RRF, condition generation on them.

  \item Verification Loop:  
    As you generate (e.g., by sentence), verify or abstain if needed.

  \item Training Alignment:  
    Use a factual regularization loss, weighted by phase, to fine-tune or align the model.

  \item Self-Reflective Revision:  
    Optionally, allow the model to critique and revise its own output.
\end{enumerate}

\section{Discussion and Future Work}

\subsection{Limitations}

- The computational overhead is substantial.  
- Empirical validation of phase‑based uncertainty (\( \sigma^2_\phi \)) is needed.  
- Retrievers can introduce irrelevant or noisy context.  
- Verification models may themselves hallucinate.  
- Setting real‑time thresholds for abstention / regeneration is nontrivial.

\subsection{Future Directions}

- Study the empirical link between positional phase \( \phi_t \) and hallucination risk.  
- Develop efficient approximations to semantic entropy (e.g., low-rank kernels).  
- Explore hybrid symbolic + neural verification.  
- Learn policies for when to retrieve, abstain, or regenerate, based on uncertainty.  
- Build human-in-the-loop systems: abstention sends output + uncertainty metrics to humans.

\section{Conclusion}

Hallucinations in LLMs stem from deep probabilistic, information-theoretic, and architectural issues: compounding error, distribution mismatch, miscalibration, and epistemic ignorance. By unifying Bayesian uncertainty estimation, semantic entropy, phase modeling, and principled decoding + training strategies, we offer a mathematically coherent framework to make LLMs more reliable. While there are practical challenges, this approach sets a strong foundation for future systems that are both powerful and trustworthy.

\appendix

\section{Appendix: Additional Mathematical Formulas}

\subsection{Fourier‑Style Phase Decomposition}

Let \( \phi_t \) be the sinusoidal phase for position \( t \). We can decompose the logit \( z_t \) as:

\[
z_t = A_0 + \sum_{n=1}^{N} \left( A_n \cos(n \phi_t) + B_n \sin(n \phi_t) \right)
\]

where \( A_n, B_n \) are learned coefficients.

\subsection{Complex‑Valued Probability with Phase}

Define a complex logit \( z_i = u_i + i v_i \). Then:

\[
p(y \mid X) = \frac{ \exp(u_y + i v_y) }{ \sum_{i \in \mathcal{V}} \exp(u_i + i v_i) }
\]

Magnitude and argument:

\[
|p(y \mid X)| = \frac{ e^{u_y} }{ \sum_i e^{u_i} }, \quad \arg(p(y \mid X)) = v_y - \arg\left(\sum_i e^{u_i + i v_i}\right)
\]

\subsection{Phase‑Weighted Factual Loss}

\[
\mathcal{L}_{\text{phase}} = \mathcal{L}_{\mathrm{CE}} + \lambda_{\mathrm{fact}} \cdot \mathcal{R}_{\mathrm{fact}}(\hat y, S_{\mathrm{verifier}}) \cdot \left(1 + \sin^2(\phi_t) + \cos^2(2\phi_t)\right)
\]

\subsection{General Oscillatory Uncertainty}

A richer formulation:

\[
\sigma_{\mathrm{osc}}^2(x_t) = \sigma_{\mathrm{epi}}^2(x_t) \cdot \left(1 + \alpha \cos^2(\phi_t) + \beta \sin^2(3 \phi_t) + \kappa \tan^2(2 \phi_t)\right)
\]

\subsection{Phase-Shifted Kernel Density Matrix}

Let \( K \) be a semantic kernel among candidates, and \( W(\phi) = \mathrm{diag}(e^{i \phi_1}, \dots, e^{i \phi_N}) \). Define:

\[
\rho_\phi = \frac{1}{Z}\,W(\phi)\,K\,W(\phi)^{-1}, \quad Z = \mathrm{Tr}(W(\phi)\,K\,W(\phi)^{-1})
\]

Compute the von Neumann entropy:

\[
S(\rho_\phi) = -\mathrm{Tr}\bigl(\rho_\phi \log \rho_\phi\bigr)
\]
\nocite{*}  \clearpage
\bibliographystyle{unsrtnat}  
\bibliography{hallucination_math_refs}
\clearpage
\centering {Glossary of Mathematical Terms}
\centering
\begin{tabular}{@{} p{4cm} p{10cm} @{}}
\hline
\textbf{Term} & \textbf{Meaning / Definition} \\ \hline

Entropy (Shannon) & A measure of the average uncertainty or “surprise” in a probability distribution: how unpredictable an event is on average. \\

Self‑information (“surprisal”) & The information gained from observing a specific event \(x\): \(I(x) = -\log p(x)\). Low-probability (rare) events carry more surprisal. \\

Kullback–Leibler (KL) Divergence & A way to quantify how one probability distribution diverges from another “true” distribution:  
\(\displaystyle D_{\mathrm{KL}}(P \,\|\, Q) = \sum_x P(x) \log \frac{P(x)}{Q(x)}\). \\

Conditional Entropy & The expected uncertainty remaining in a random variable \(Y\), given that we know another variable \(X\). \\

Calibration / Expected Calibration Error (ECE) & A metric for how well a model’s predicted probabilities align with its actual accuracy; large ECE means the model is “confidently wrong” often. \\

Epistemic Uncertainty & Uncertainty in the model’s predictions due to lack of knowledge (e.g., model parameters or out-of-domain data). Often measured via Bayesian techniques. \\

Monte Carlo Dropout & A method to approximate Bayesian uncertainty by performing multiple forward passes with dropout turned on at inference and then measuring variation. \\

Von Neumann Entropy & A quantum‑like entropy for a density matrix \(\rho\): \(S(\rho) = -\mathrm{Tr}(\rho \log \rho)\). Measures “mixedness” or uncertainty in a quantum‑state analogy. \\
Kernel Language Entropy (KLE) & Semantic uncertainty: build a kernel on candidate continuations, convert it into a density matrix, and compute its von Neumann entropy to assess semantic diversity. \\

Positional Phase (in Transformers) & In sinusoidal positional embeddings, each token has a “phase” value \(\phi_t\); this can be used to tie uncertainty or regularization to token position. \\

Fourier Decomposition of Logits & Expressing a logit \(z_t\) as a sum of sinusoids of the phase:  
\[
z_t = A_0 + \sum_{n=1}^N \bigl(A_n \cos(n \phi_t) + B_n \sin(n \phi_t)\bigr)
\] 
to capture periodic positional effects. \\

Phase‑Modulated Variance / Uncertainty & Uncertainty that depends on the positional phase, e.g.:  
\[
\sigma^2_\phi(x_t) = \sigma^2_{\text{base}}(x_t) \cdot (1 + \gamma \sin^2(\phi_t))
\]  
So some positions may be more “risky.” \\

Contrastive Decoding Score & A scoring function in decoding that penalizes tokens favored by a baseline model and optionally adds a phase bias:  
\[
\mathrm{Score}_{CD}(x_t) = \log P_{\text{full}}(x_t) - \lambda \log P_{\text{baseline}}(x_t) + \eta \sin(\phi_t)
\] \\

Reciprocal Rank Fusion (RRF) & A method to combine multiple retrieval rankings:  
\[
\mathrm{RRF}(r) = \sum_{q \in Q} \frac{1}{k + \mathrm{rank}_q(r)}
\] \\

Real‑Time Rectification / Abstention & A strategy: generate → verify (via retrieval or critic) → if the content is not trustworthy, either refuse (abstain) or regenerate. \\

Self‑Evaluation / Self‑Alignment & A loop where the model evaluates its own outputs (e.g., for factuality), then revises or abstains accordingly. \\ \hline

\end{tabular} 
 \begin{table}[htbp]
  \centering
  \caption{Glossary of mathematical and information‑theoretic terms used in this paper.}
  \label{tab:glossary}
\end{table}

\end{document}